\documentclass[letterpaper, twocolumn, 10pt]{article}
\usepackage{graphicx}
\def\BibTeX{{\rm B\kern-.05em{\sc i\kern-.025em b}\kern-.08em
    T\kern-.1667em\lower.7ex\hbox{E}\kern-.125emX}}
\usepackage{svg}
\usepackage[round]{natbib}

\usepackage{balance}

\usepackage[none]{comment} % Highlight edits and show comments

\begin{document} 

\title{Ground Compliance Improves Retention of Visual Feedback-Based Propulsion Training for Gait Rehabilitation}%On the After-Effects of Intentionally Increasing Push-Off Forces on Compliant Terrain: \\A Bio-Feedback Approach to Gait Rehabilitation \P{needs to change}}

\author{Bradley Hobbs, and Panagiotis Artemiadis
\thanks{*This material is based upon work supported by the National Science Foundation under Grants No. \#2020009, \#2015786, \#2025797, \#2018905 and \#2415093 and the National Institutes of Health Grant No. 1R01HD111071-01. }
\thanks{Bradley Hobbs and Panagiotis Artemiadis are with the Mechanical Engineering Department, at the University of Delaware, Newark, DE 19716, USA.
        {\tt\small bwh@udel.edu, partem@udel.edu}}
\thanks{Corrresponding author: partem@udel.edu}
}

\date{}
\maketitle

\begin{abstract}
%Visual feedback is a crucial component of coordinating goal-directed walking strategies. In gait training, visual feedback has shown to be a useful tool for improving kinetic outcomes such as push-off force (POF), which is associated with favorable recovery outcomes following motor impairment such as stroke. These positive outcomes in gait kinetics can be further studied by inducing kinetic changes in the walking environment. The goal of this work is to investigate ground compliance as a method for increasing POF and compare it to POF strategies when using visual force feedback alone. This study involved 10 healthy participants who received real-time visual feedback of ground reaction forces while walking, with some also experiencing ground compliance changes using a custom split-belt treadmill, to explore strategies for gait rehabilitation. The main finding is that intentional increases in propulsive ground reaction forces (POF) were sustained post-intervention -especially in the compliant ground group - with lasting after-effects also observed in muscle activity and joint kinematics. Therefore, it is shown that adding ground compliance to visual feedback-based gait training enables robust learning of natural strategies to increase propulsion force. This is significant because it reveals how visual and proprioceptive systems coordinate during gait adaptation and supports the potential for using compliant terrain in long-term rehabilitation targeting propulsion deficits.

\textbf{Purpose:} This study investigates whether adding ground compliance to visual feedback (VF) gait training is more effective at increasing push-off force (POF) compared to using VF alone, with implications for gait rehabilitation.

\noindent\textbf{Design/methodology/approach:} Ten healthy participants walked on a custom split-belt treadmill. All participants received real-time visual feedback of their ground reaction forces. One group also experienced changes in ground compliance, while a control group received only visual feedback.

\noindent\textbf{Findings:} Intentional increases in propulsive ground reaction forces (POF) were successfully achieved and sustained post-intervention, especially in the group that experienced ground compliance. This group also demonstrated lasting after-effects in muscle activity and joint kinematics, indicating a more robust learning of natural strategies to increase propulsion.

\noindent\textbf{Originality:} This work demonstrates how visual and proprioceptive systems coordinate during gait adaptation. It uniquely shows that combining ground compliance with visual feedback enhances the learning of propulsive forces, supporting the potential use of compliant terrain in long-term rehabilitation targeting propulsion deficits, such as those following a stroke.


\noindent\textbf{Article Classification:} Research Article
\end{abstract}

%\begin{IEEEkeywords}
%robotics, gait, bio-feedback, rehabilitation
%\end{IEEEkeywords}

\section{Introduction}
% VF 
In cases of brain lesions, loss of lower limb muscle control is common \citep{verma2012}, leading to the necessity of complex gait retraining strategies \citep{anson2013,franz2016,awad2014,awad2020,krishnamoorthy2008,liu2021}. Increasing a patient's muscle activation and coordination is often a primary goal \citep{dwyer2010,reiman2012,akdougan2011}. It is also clear that joint angle trajectories should be noted during training to ensure proper gait patterns are being developed in training \citep{vallery2008,ji2008,ensink2025}. Many motor therapy techniques incorporate biofeedback \citep{huang2006,basmajian1981,richards2017,lewek2012,nelson2007} such as live visual feedback (VF) \citep{sigrist2013,giggins2013} using the participant's own joint angles \citep{banz2008} or toe clearance \citep{banala2010novel,banala2007} to decrease incidence of drop-foot in participants with hemiplegia \citep{hsu2019,lemoyne2008,dingwell1996}. 

% POF 
Force bio-feedback has been used for enhancing anterior-directed push-off forces \citep{schenck2017,genthe2018,franz2014,herrero2021,white1997}. These forces are a primary variable of concern \citep{spencer2021,vc2023aftereffects,kim2024gait,roelker2019,zelik2016,hsiao2015} since gait kinetics can relate multiple therapy outcomes \citep{mirelman2010,chen2007,hesse1994,chen2003}. Indeed, the 3-dimensional push-off force is a conjoining measure of not only the performance of ankle musculature \citep{kameyama1990,schenck2019}, but also in the kinetic chain of the entire lower body as a whole \citep{dugan2005}, with the anterior-posterior component most responsible for forward-directed propulsive forces \citep{balasubramanian2007}, and the purely vertical component displaying the largest magnitude \citep{perry1992}. Previous work has shown that participants given feedback of only vertical component of ground reaction forces are able to increase POF after training to intentionally increase POF \citep{vf1}. 

% low stiffness
Modifications to the expected ground reaction force profile during loading response and terminal stance can disrupt sensory pathways to the brain \citep{perry1992,preusser2015,kim2015}. This is one reason why investigations of the past decade have included the study of compliant (low-stiffness) environments, which have the potential to perturb multiple neural pathways simultaneously. For example, proprioception is altered as the foot descends below the typical ground level during force application, necessitating postural adjustments \citep{tuthill2018,rand2018,ting2015}. Inter-limb coordination mechanisms are also challenged when using asymmetric compliant terrain changes \citep{handelzalts2019,skidmore2016a,skidmore2016effect}, since current research posits that asymmetric gait patterns may be corrected through asymmetric interventions. These multifaceted stiffness changes aim to stimulate neural plasticity in a unique manner, potentially improving outcome measures on the paretic side. 

Compliant environments are typically emulated by walking overground on varied walking surfaces \citep{xie2021} similar to deformable ground found in nature. In order to localize and better control the compliance of the walking surface in a laboratory environment, a robotic split-belt treadmill, Variable Stiffness Treadmill (VST), was developed to provide highly repeatable and controllable compliant surfaces \citep{barkan2014}. Notable findings include significant evoked brain activity through low stiffness perturbations \citep{skidmore2016a}, substantial contralateral muscle activation \citep{skidmore2016a}, and significant after-effects through long-term use \citep{chambers2023}. More recently, VST 2 was developed \citep{chambershobbsVST2} and is shown in Fig. \ref{fig:setup}. VST 2  contains numerous improvements to VST, including independent stiffness and speed for both belts, linear vertical deflection instead of angular, wider range of stiffness values, quadrupled walking area and center of pressure sensing area, half of the ground height, and quieter operation \citep{chambershobbsVST2}. At the time of conducting this study, VST 2 is the only treadmill able to adjust the compliance of the walking surface.

% building upon prev works
This study aims to evaluate the effectiveness of combining reduced ground stiffness with real-time, continuous visual feedback of vertical force for gait training, specifically targeting an increase in push-off force after-effects. We propose a novel protocol and improve upon the current literature in several key ways. First, the present study extends previous work by including compliant terrain during the intervention and studying the associated after-effects. Second, a control group is used to isolate the effects of using visual feedback alone, providing a baseline for comparison. Much of the previous work that has shown promising kinetic results using visual feedback used within-participant study designs. Third, vertical ground reaction forces are reported, since previous work primarily focuses on anterior-posterior forces during push-off. Fourth, measurements of muscle activity around the hip and knee are reported, as well as joint kinematics while push-off occurs. It is imperative that these are assessed in order to determine fully if visual force bio-feedback can be used to increase push-off force, especially before using this method in conjunction with interventions that utilize more complex forms of robot-assisted therapy \citep{ck2023pros}.

% Hypotheses

In summary, this study is based on the hypothesis that integrating low-stiffness compliant terrain with visual feedback (VF) gait training yields greater and more persistent increases in push-off force (POF) compared to training with VF alone. We test this using a novel protocol that evaluates sustained push-off after-effects from this combined approach, offering a comprehensive kinetic and muscular analysis critical for advancing complex robot-assisted gait therapies.

%We hypothesize that incorporating low stiffness into current gait training methods can lead to better outcomes compared to training methods that do not utilize low stiffness. Further, it is hypothesized that even individuals trained to increase peak push-off force consistently can still benefit from walking over low stiffnesses. This study introduces a novel protocol that uniquely integrates compliant terrain with real-time visual feedback to evaluate sustained push-off after-effects, offering a comprehensive kinetic and muscular analysis critical for advancing complex robot-assisted gait therapies.

%This study is the first to provide muscle activity and joint kinematics of a separate group walking on VST 2 set to rigid that support implementing compliant terrain in conjunction with visual force bio-feedback in rehabilitation protocols.\Pc{the last sentence is not strong, and probably redundant. Replace with a sentence summarizing the significance of the problem instead.}

\section{Methods}
\subsection{Experimental Setup}
\begin{figure}[t!]
  \begin{center}
    \includegraphics[width=0.97\columnwidth,trim={70 0 90 0},clip]{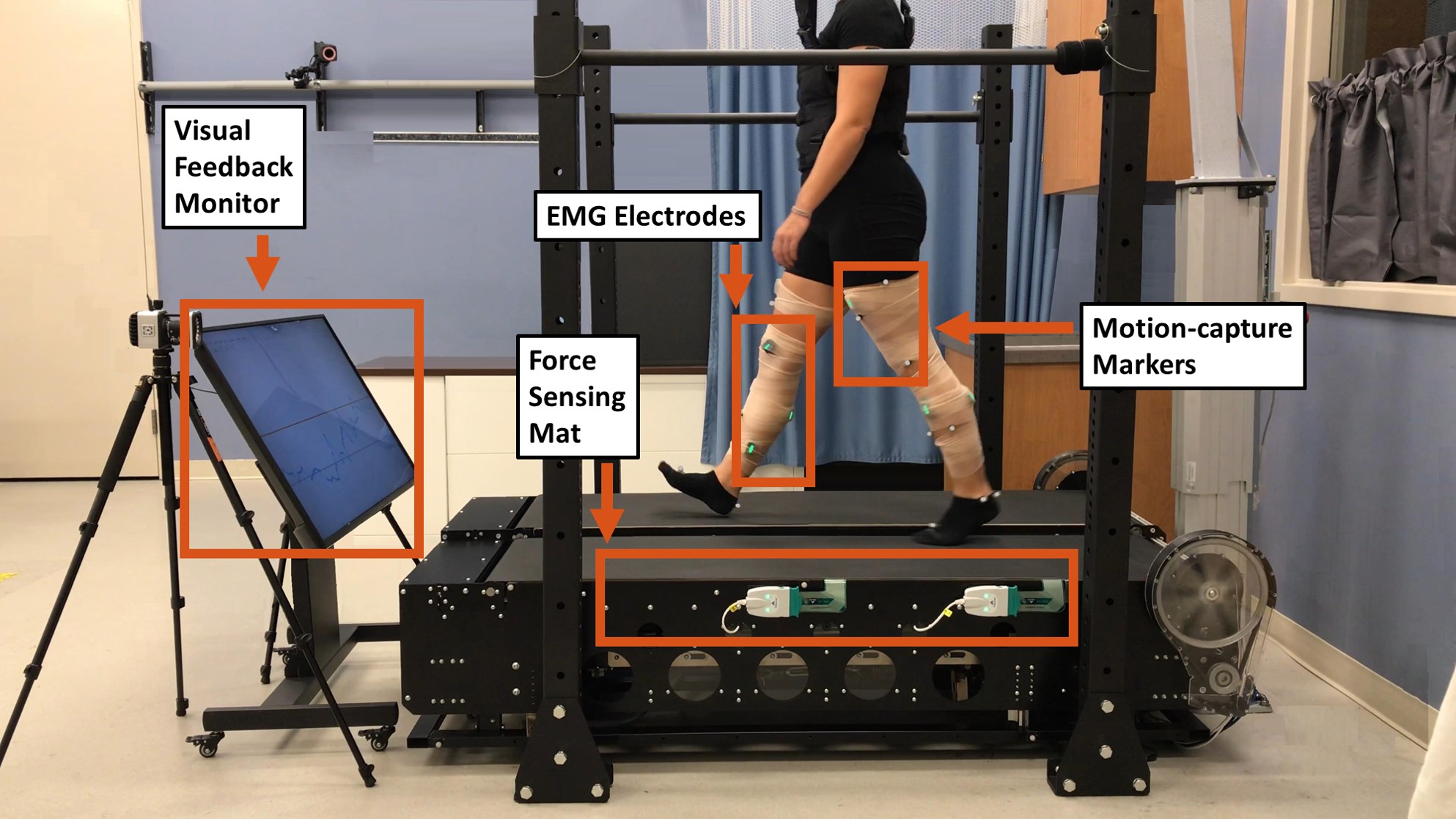} 
    \caption{Experimental setup showing a representative participant wearing reflective markers and EMG electrodes on the legs while walking on VST 2 equipped with the force sensing mats, and observing the visual feedback monitor.}
    \label{fig:setup}
  \end{center}
\end{figure}

% VST 2
For this experiment, a unique instrumented robotic split-belt treadmill, the Variable Stiffness Treadmill 2 (VST 2) was used, as shown in Fig. \ref{fig:setup}. This device has the ability to change the walking surface stiffness on each belt side independently due to a unique mechanism that includes a spring-attached lever system. The position of this lever's fulcrum is controlled through a lead screw-bound carriage that can move with high resolution to different positions to quickly change the relative stiffness felt by the user. More specifics of this design are detailed in previous works \citep{chambershobbsVST2}. For this study, the left belt lowered stiffness to 25~$kN/m$, while the right belt remained rigid. Due to the nature of platform height changes, all participants wore a safety harness (Litegait Inc.) capable of supporting the entire body weight of the participant if needed, with the straps adjusted to retain slack even through the lowest platform deflection during a gait cycle with lowered stiffness.

% Forcemats
Vertical ground reaction forces for both sides of the VST 2 were captured using four TekScan 3140 Medical Sensors (TekScan Inc.). These flat mat-like sensors were located under the treadmill belts and in total contain 8,448 individual force-sensing cells in a grid pattern, with a resolution of 1 cell per centimeter and a recording frequency of 64~$Hz$. This allows the center of pressure (COP) location to be precisely determined in real time. The raw data were filtered using techniques detailed in previous work \citep{ck2024grf}. During the experiment, the data were streamed to a computer using software in C\# for storing and for utilization in the visual feedback. The visual feedback was shown on a 90~$cm$ high-resolution monitor that was placed directly in front of VST 2 (see Fig. \ref{fig:setup}). Stored force mat data were synchronized with both EMG and motion capture data through a data acquisition device (National Instruments Corporation) that receives a trigger signal from the Vicon Nexus software (Vicon Motion Systems Ltd). In order to decrease interference with normal gait mechanics and head position, the monitor was also placed so that the gaze of the participant was slightly downward.

% Mocap
For capturing lower-limb joint kinematics, a motion capture camera system (Vicon Motion Systems Ltd.) was used in conjunction with 20 reflective markers placed at strategic positions. These markers' positions were streamed via 8 high-speed wide-angle cameras at a frequency of 100~$Hz$. The live marker positions correspond to specific anatomical positions, including the participant's heel, which was used for real-time detection of the heel-strike events that were used to define the start of a gait cycle \citep{ck2021real}. Using all 20 markers, the participant's bone position and orientation were calculated, giving accurate angle values for all lower-body joints. 

% EMG
Muscle activity was recorded using the Delsys Trigno wireless surface electromyography (EMG) system (Delsys Inc.) at 2000~$Hz$. A total of 16 electrodes were used, with the recorded muscles corresponding to the following: Tibialis Anterior (TA), Gastrocnemius Medial (GAM), Gastrocnemius Lateral (GAL), Soleus (SOL), Vastus Medialis (VM), Rectus Femoris (RF), Biceps Femoris (BF), and Semitendinosus (ST) for both the left and right legs. Because this is the first study to report vertical ground reaction forces and kinematics using a protocol for increased POF, the largest and most active muscles contributing to the movement patterns involved were included. Temporal synchronization between all data collection devices was achieved through a digital trigger unit and Vicon Nexus 2.12 (Vicon Motion Systems Ltd.). All participants had EMG electrodes placed according to current guidelines \citep{David2024}, including shaving excess hair, wiping the skin with alcohol, securing with double-sided tape to the skin and elastic wrap around the limb to reduce noise due to movement.

\subsection{Experimental Protocol}
% participants,IRB
The protocol for this study was approved by the University of Delaware Institutional Review Board (IRB ID\#: 1544521-2), with informed consent given by all participants. participant inclusion criteria include being free from any injuries affecting the lower limbs, gait abnormalities, and attention difficulties. A total of ten participants (5 male, 5 female; age = 25.8 $\pm$ 3.1 years; body mass = 72.6 $\pm$ 7.9~$Kg$; height = 177.4 $\pm$ 9.5~$cm$) participated in this study, with each participant having the right leg dominant. All participants walked without shoes. The entire experiment was separated into two halves: the Training half, and the Trial half. 

\begin{figure}[t!!]
  \begin{center}
    \includegraphics[width=0.94\columnwidth,trim={30 30 47 20},clip]{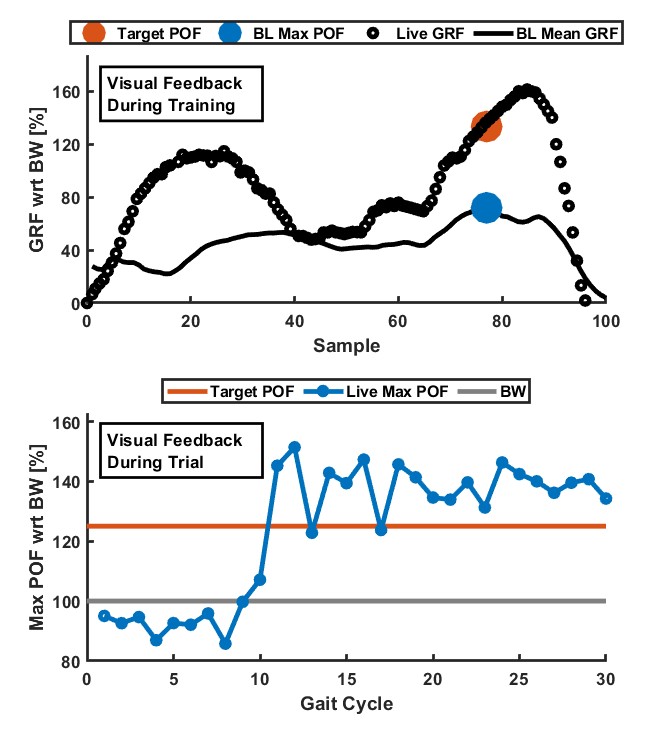} 
    \caption{Close look of the visual feedback shown to the participants with training feedback shown on top and trial feedback shown on the bottom. The axes units, labels, or legends are not shown to the participants, but are included in the figure for clarity, with Baseline (BL) values given with respect to (wrt) percent bodyweight (BW).}
    \label{fig:vf}
    \vspace{-0.5cm}
  \end{center}
\end{figure}

\begin{figure*}[t!]
  \begin{center}
    \includegraphics[width=0.97\textwidth,trim={12 12 0 8},clip]{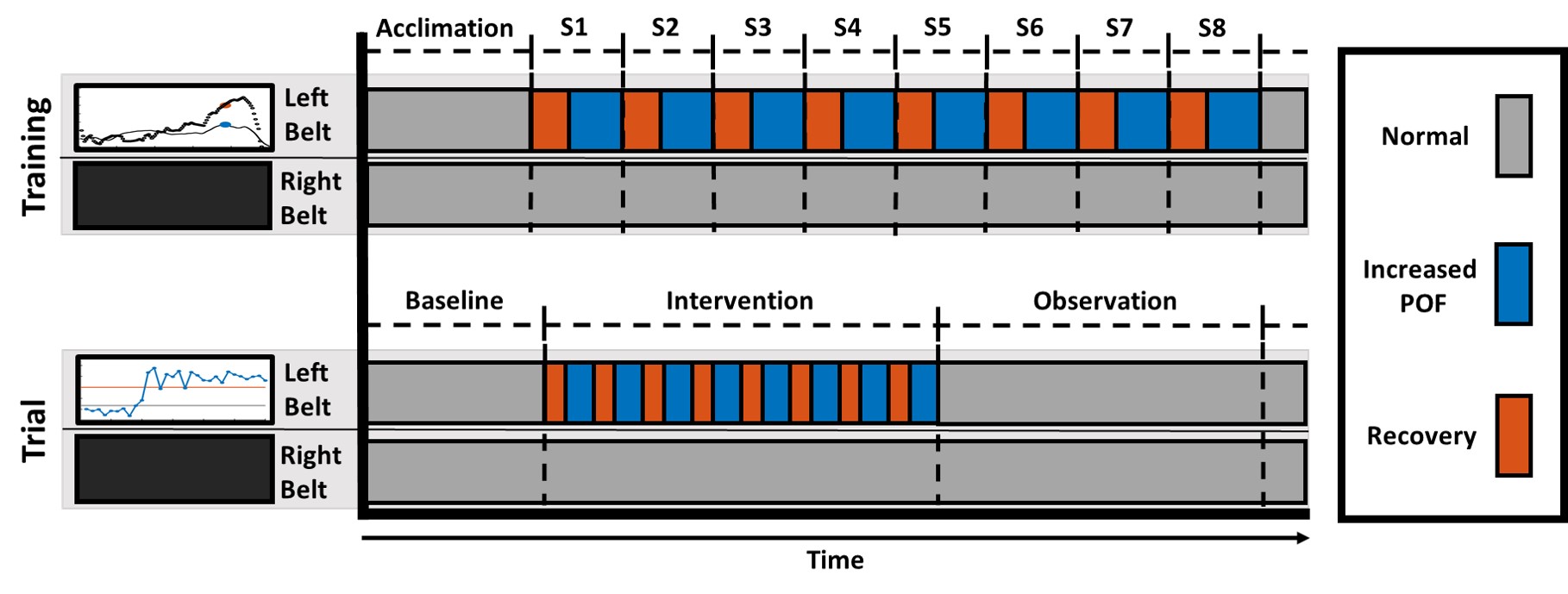} 
    \caption{For each group, the visual feedback used is shown on the left, with the relative number of gait cycles in each trial half proportioned through time, and for each belt individually. Each group received the same visual feedback and was equally trained to increase POF.}
    \label{fig:protocol}
    \vspace{-0.5cm}
  \end{center}
\end{figure*}

\subsubsection{Training Protocol}
% Training Acclimation
For the first half of the study, all participants were instructed to walk normally on VST 2 at a constant speed of 90~$cm/s$, and for a duration of around 100 gait cycles, comprising the Acclimation section of the Training. After the participants were comfortable with walking on VST 2, the belts were stopped in order to go over further instructions. participants were then instructed to closely observe the VST 2 monitor at all times when walking, and that the monitor will provide a visual indication as to the amount of vertical force being exerted on the left belt while walking (see Fig. \ref{fig:vf}). 

% Training Baseline
For the first ten gait cycles, the participants were shown a blank screen. Then, a baseline average of each individual participant's ground reaction force (GRF) profile from the left leg only was shown on screen for the rest of the training half as a static image through a live updating Matlab figure window (MathWorks, Inc.). As the participant walks, each of the ground reaction force magnitude samples was cumulatively displayed as a black point on the screen, with the x and y coordinate positions proportional to the current time instant's gait cycle phase, and proportional to the amount of ground reaction force, respectively. In other words, the participant was able to see the vertical force exerted on the treadmill by the left leg in real time and relative to the calculated baseline as a function of time. A baseline (BL) dot was shown to represent the peak push-off force (POF) for the averaged baseline profile, while a target dot represents 125$\%$ of the individual's body weight (BW), since a baseline value was unknown before the start of this study. This value was chosen experimentally and based on previous work  \citep{vf1} to induce a meaningful difficulty level, while remaining attainable, thereby having the desired effect of increasing the motivation of the participant, without frustration. All these are shown in Fig. \ref{fig:vf} on the top.

% Training Target
The participants first walked for 10 gait cycles under normal conditions, which constitutes the baseline (BL) phase. This was immediately followed by 15 gait cycles, where participants were instructed to increase the peak force at push-off against the treadmill belt while monitoring their performance on the monitor. This phase is referred to as the increased push-off force (POF) phase. This visual feedback's purpose was to give as much relevant information as possible to allow the participants to adjust the POF strategy in real time. These 25 gait cycles form a single section of the training half, with eight total sections of training (200 gait cycles) given to all participants after acclimation (see Fig. \ref{fig:protocol}). Participants received standardized verbal cues to help them understand and increase POF, with success defined as reaching the on-screen target in at least 50\% of a training section. The target was framed as a binary goal, either reached or not, to encourage consistent effort while minimizing the risk of exaggerated or unnatural gait patterns during the intervention phase.

% Training additional rounds
If a participant didn't succeed within eight training sections, they rested for up to five minutes before repeating another round, continuing until success was achieved. Most required one or two rounds; two needed three. We ensured all participants received equal training levels before proceeding, followed by a five-minute rest to prevent fatigue.

\subsubsection{Trial Protocol}
% Trial BL vs OB
The Trial half of the study consists of 500 gait cycles separated into three phases: baseline, intervention, and observation (see Fig. \ref{fig:protocol}). The baseline phase consists of 100 gait cycles with both platforms set to rigid and serves as the participant's normal walking gait for all analysis. The observation phase operates under the same conditions as the baseline, but lasts for 200 gait cycles in order to give ample time to observe after-effects from the intervention phase. The purpose of this phase was to be identical to the baseline in protocol, so that changes lasting from the intervention phase can be measured against that before the intervention. The intervention phase immediately follows the baseline phase and occurs before the observation. In the intervention phase, participants walk for 200 gait cycles, with all participants completing exactly one more round of eight sections, structured exactly as in training (see Fig. \ref{fig:protocol}). All participants were given visual feedback that was adjusted to suit the goals of the Trial half.

% Trial intervention
The visual feedback used in the Trial intervention phase for all participants was a simpler version of the training visuals, because the participants were already familiar with the task, and do not need as much detailed information. As shown in Fig. \ref{fig:vf} on the bottom, a target was shown, but as a solid line across the screen. Another line was shown representing the individual's body weight, in order to have a ground reference for scale. The live feedback representation shown in this visual, however, was only the POF of the current gait cycle, which was calculated and displayed back within five milliseconds (ms) after the left toe-off (LTO) occurs. Specifically, the participant was shown a single new point for each gait cycle, giving POF performance feedback relative to the target at the end of each stance phase of the gait cycle. This point was sustained on the screen for several gait cycles, which allows the participant to visually gauge effort relative to previous gait cycles, which has never been used for this type of intervention protocol.

% Trial low stiffness
It was during the intervention phase that the participants were dissected into two groups: the Visual Feedback with Low-Stiffness (VF-S) group, and the Visual Feedback with rigid (VF-R) group, serving as a control. The VF-S group has the left platform of VST 2 set to reduced stiffness of 25~$kN/m$ for only the duration in which POF was intentionally increased (15 gait cycles), and then set back to rigid for the next 10 gait cycles. Because this low-stiffness environment has not been experienced by the participants previously, the VF-S group had to adapt to this change in real-time. All participants still had eight attempts to reach the target shown on the screen, with the VF-S group requiring low-stiffness walking on the left side. In order to duplicate any auditory feedback experienced by the VF-S group due to the stiffness changes, the VF-R group protocol includes having the VSM move out during the swing phase, but return to rigid values before heel-strike.
% \Pc{syntax is strange, and also, how did you do that without changing the effective stiffness?}. I changed the stiffness from 1000 to 999 kn/m which is enough to hear it very briefly, but not enough to deflect any differently

\subsection{Data Pooling and Pre-Processing}
% All data
Using the same method for each participant, all data processing was completed individually, with further individual separation between participant groups, experiment half, and walking phase. Then, for each participant individually, data outliers were removed through an advanced systematic outlier detection algorithm described in \citep{hobbs2022} for all kinematic, kinetic, and muscle data. The entirety of all phases is interpreted. 

% Figures
For analysis of POF throughout the trial half, all participants were pooled to draw conclusions about the intervention as a whole with regards to the POF, which was the primary focus of the present work. For secondary analysis, two participants were pulled from each group to represent the average of each group, with an equal portion of participants performing higher and lower in terms of POF. This allows direct comparison between groups without normalizing further data to an average. 

% Stats
For all data, statistical tests were run using the Wilcoxon rank-sum test with $\alpha = 0.05$. This test was used due to the low sample sizes and the data distribution. In visualizations used in this paper, when two datasets were found to be statistically significant, an asterisk is used in conjunction with a line connecting the two datasets being compared. The two datasets being tested in this study were always a two-tailed comparison between data from the observation phase and the baseline phase. For POF data, all gait cycles in the baseline phase were used, with the observation phase split into two halves and evaluated separately due to the transient nature of the observation phase while still accounting for having a sufficient amount of data for comparison. For all other data, the difference between observation and baseline were used for each group. Each half of the observation phase was tested for significance relative to the entire baseline phase. For boxplots, the method for each comparison is indicated at the top of its respective significance line, with the directionality of hypothesis testing indicated by arrow symbols. An upward arrow denotes a right-tailed test, where the alternative hypothesis posits that the observation phase exceeds the baseline phase, while a downward arrow signifies a left-tailed test, examining whether the observation phase was significantly lower than the baseline phase. The absence of an arrow indicates that a two-tailed test was implemented, evaluating significant differences in either direction between the observation and baseline phases.

% POF, GRF
The POF for the entire Trial half was compiled into the corresponding group and normalized with respect to each participant's bodyweight and subtracted from the average POF of the respective baseline phase to show the intervention and observation phases for each group relative to a common baseline for visual comparison. Statistical testing was used to determine if the POF from the VF-S group was significantly higher than the VF-R group. GRF data were normalized to the participant's bodyweight and filtered using methods from previous work \citep{vf1}. Both the left and right GRF profiles were shown for each group, with only the stance phase displayed. For the left leg, the data spans between left heel-strike (LHS) and left toe-off (LTO), while the right leg spans between right heel-strike (RHS) and right toe-off (RTO). 

% Mocap, EMG
Motion capture data were used to calculate 3D joint angles using the Vicon Plug-in-Gait model (Vicon Motion Systems Ltd.), which uses 3D marker positions and each participant's anatomical parameters. EMG data were filtered using a fourth-order Butterworth band-pass filter using cut-off frequencies at 30 and 300~$Hz$. After full-wave rectification, a 200 data point moving average calculates the muscle activity envelope, and a fourth-order, 5~$Hz$ lowpass filter was used. Data were then normalized with respect to the highest recorded value during the experiment, or experiment maximum (EM). This normalization method consists of the peak dynamic method \citep{Burden2003}, which reduces inter-participant variability when individual muscle effort comparisons were not needed. For both EMG and kinematic data, dynamic time warping was performed using standard spline to the highest number of samples available in a given participant, and spans between left heel-strikes, which were calculated offline using the heel marker \citep{ck2021off}. 

\begin{figure}[t!]
  \begin{center}
    \includegraphics[width=0.97\columnwidth,trim={40 45 78 49},clip]{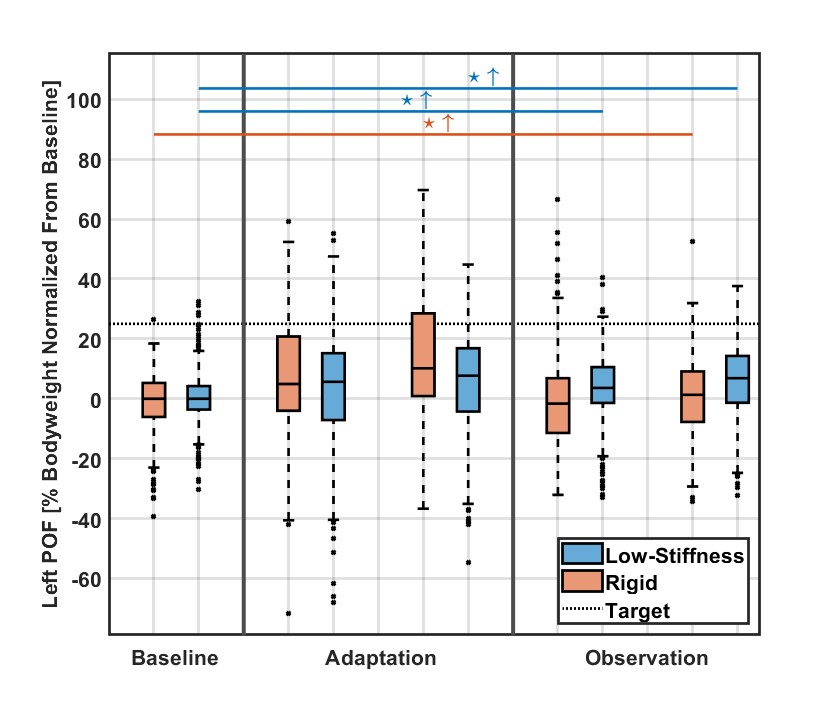}
    \caption{Peak vertical POF values for all participants in each group, normalized to the average POF during the last baseline (BL) phase for each participant individually. Colored bars indicate a significant increase between each half of the observation phase and the entire baseline phase.}
    \label{fig:pof}
    \vspace{-0.5cm}
  \end{center}
\end{figure}

\subsection{Data Post-Processing}

% data analyzed all participants and representative
For POF data showing after-effects, a cumulative normalized combination of all 10 participants that participated in this study was analyzed. Further in-depth analysis is shown for a representative participant from each group so that data does not have to be normalized for each participant. All POF data is shown with all participant data pooled together. The representative participants were chosen to illustrate typical group performance, selected for their proximity to the group mean rather than as high-performing outliers

% all GCs used. early and late
For statistical analysis, all gait cycles are used. For boxplots, all gait cycles are used and displayed, but with the intervention and observation phase, each was separated into two equal parts to signify the early portion and the late portion. The baseline phase always references the complete set of baseline data, both for statistical analysis and display. In the boxplots, the line denoting the participant's target during the intervention phase is shown. Participants only saw this target during the entire training half and only during the intervention portion of the trial half.

% figure colors
All of the following figures retain the same color for each corresponding group. Results pertaining to the low-stiffness group are always shown with a shade of blue, while outcomes pertaining to the rigid group are always shown with a shade of orange.

% study goal
Finally, this study focuses on the observation phase compared to baseline; the intervention phase is not analyzed here, as it was previously studied using visual biofeedback during rigid walking \citep{vf1}.

\section{Results}
\begin{figure}[t]
  \begin{center}
    \includegraphics[width=0.97\columnwidth,trim={80 0 85 5},clip]{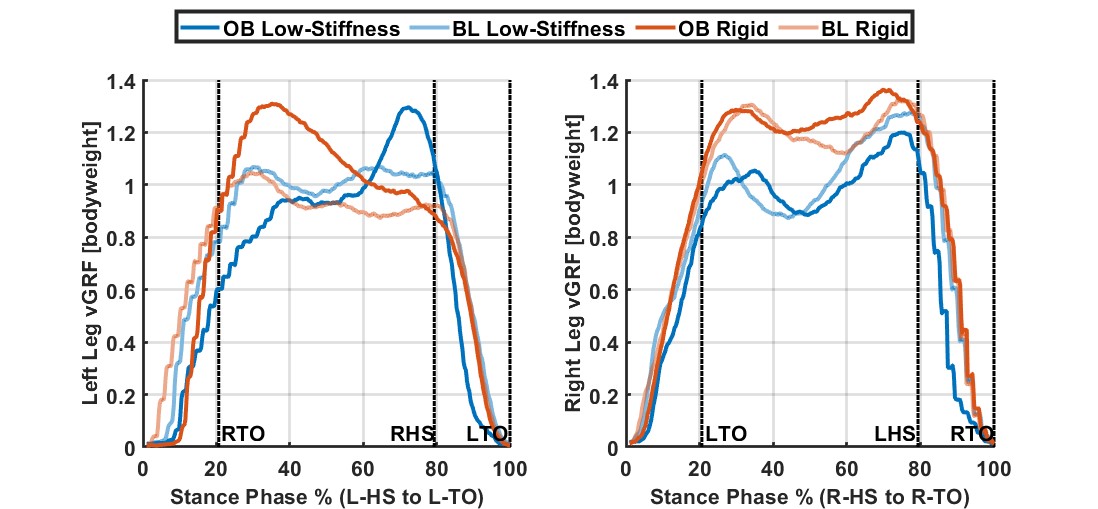} 
    \caption{GRF profile for both legs during the final trial half for a representative participant from each group. All force values are normalized with respect to body weight (BW) in order to accurately show comparisons. Blue corresponds to the low stiffness group, while orange denotes the rigid group. For each group, the darker color represents the observation (OB) phase, with the baseline (BL) phase given by the lighter color .}
    \label{fig:grf}
    \vspace{-0.5cm}
  \end{center}
\end{figure}

\begin{figure*}[t]
  \begin{center}
      \includegraphics[width=0.97\textwidth,trim={165 0 150 5},clip]{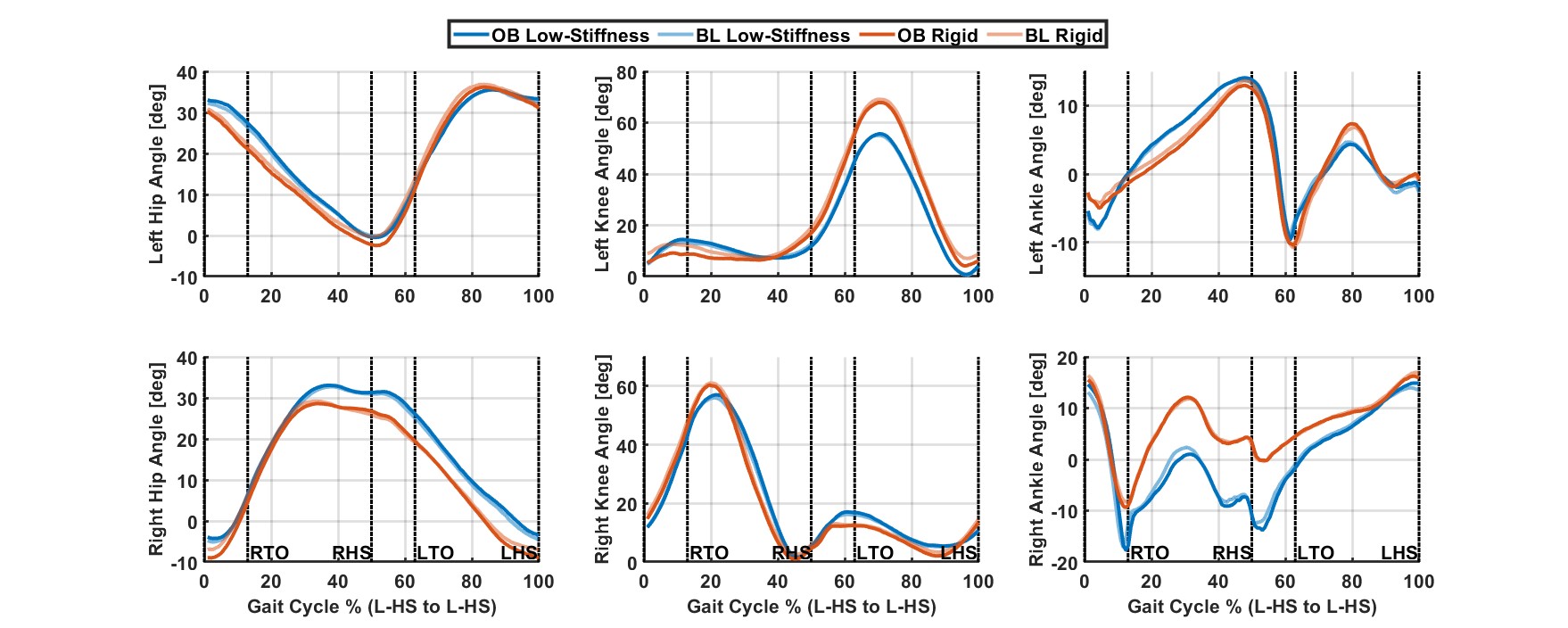} 
      \caption{Results showing left and right joint angles of the participants representing the average for the trial in each group.}
      \label{fig:kin}
      \vspace{-0.5cm}
  \end{center}
\end{figure*} 

\subsection{Intervention (Adaptation) Phase}
% trial intervention phase--------------------------------------------------------------------
All participants met or exceeded the POF target in at least 50\% of the final training section and retained increased POF into the adaptation phase, often surpassing prior maximums. This was accompanied by a 14\% increase in stance-phase duration, and all groups consistently exceeded the 125\% bodyweight POF target, aligning with previous findings \citep{vf1}. Moreover, the group undergoing rigid conditions (control group) seemed to have a small increase in peak POF, suggesting that the lowered stiffness may make it more difficult to increase POF during an intervention phase. Further work analyzing effort with each of these conditions could help elucidate if lowered stiffness requires more effort to increase POF, if the effect was due to specific strategies used, or even if the effect was due to the nature of the forcemat sensor.

% trial OB vs. BL:--------------------------------------------------------------------
\subsection{Observation Phase vs. Baseline Phase}
% kinetics pof--------------------------------------------------------------------
Findings of the current study suggest that all participants, on average, had a significantly retained increase in POF after walking under a low stiffness condition when compared to participants who were walking on rigid ($p < 0.05$). This occurs throughout the entire observation phase, without any marked reduction towards the end of the trial (see Fig. \ref{fig:pof}). For participants who experienced rigid walking during the intervention phase, the increase in POF for the observation phase was only realized for the last half of the observation phase, with the resulting POF having a much lower magnitude.

% kinetics grf LEFT --------------------------------------------------------------------
Analysis of the ground reaction force (GRF) profiles for the VF-S group revealed an approximate 20\% increase in left leg stance-phase duration, which coincided with the significant increase in left push-off force (POF) from baseline. In contrast, no significant changes were observed in the right leg GRF profile for this group.

\begin{figure*}[t]
    \begin{center}
      \includegraphics[width=0.97\textwidth,trim={195 30 165 25},clip]{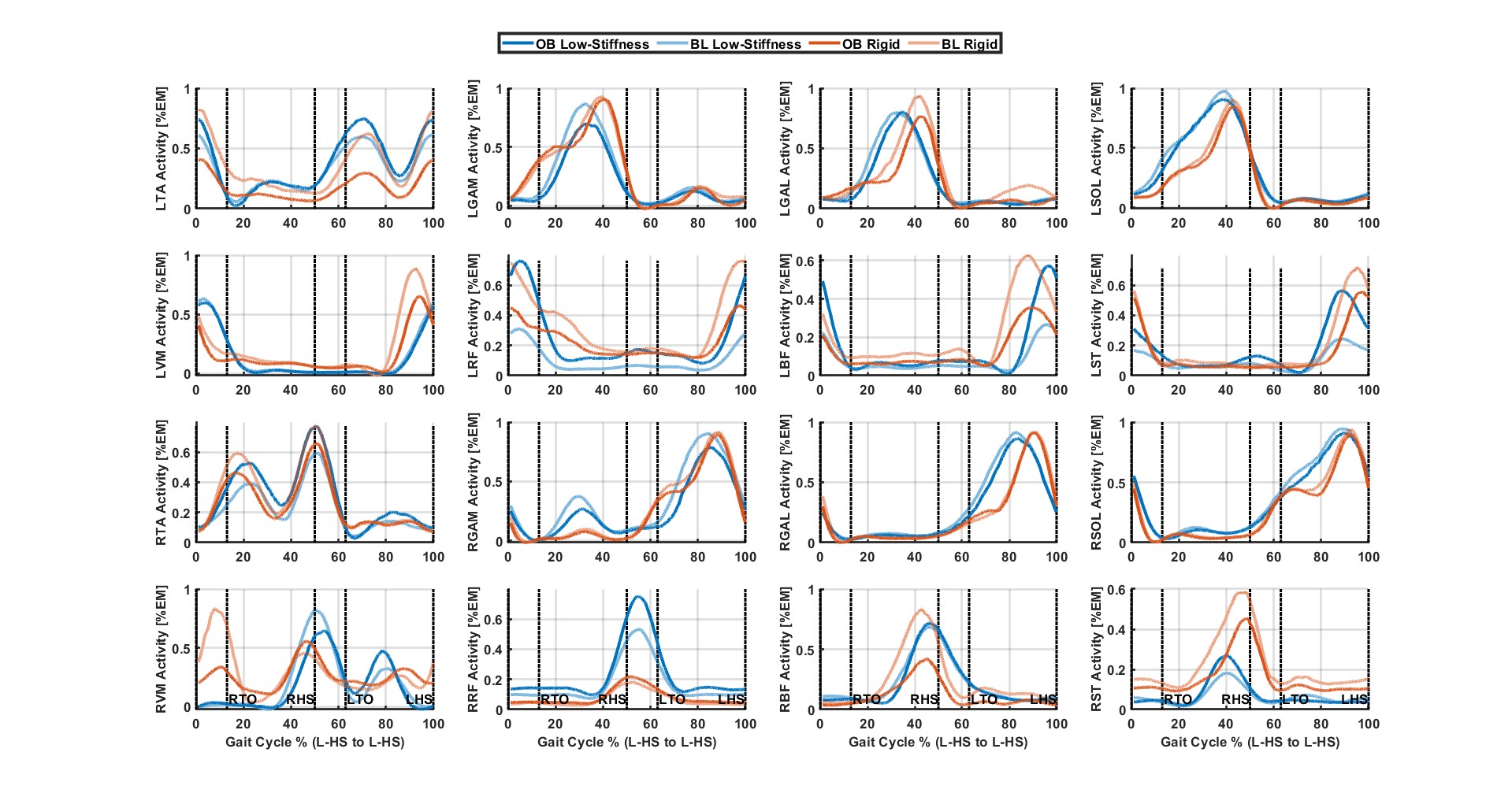} 
      \caption{Resulting muscle activity recorded for both left and right legs of the participants representing the average for the trial in each group. The units are normalized to the experiment maximum (EM).}
      \label{fig:emg}
      \vspace{-0.5cm}
    \end{center}
  \end{figure*}

% In the preparation leading up to increasing POF
Just after the left heel-strike, a knee flexion increase of 2\% was seen in the left leg of the VF-S group, while the VF-R group saw a decrease of 5\% ($p < 0.05$) (see Fig. \ref{fig:kin}). At the same time, the left ankle has an increase in peak ankle dorsiflexion of 2\% at heel-strike for the low stiffness group, while the rigid group has a decrease of 5\% ($p < 0.05$). These changes were also coupled with a remarkable change in left rectus femoris activity, with a 45\% ($p < 0.05$) increase for the VF-S group, and a 30\% ($p < 0.001$) decrease in the VF-R group (see Fig. \ref{fig:emg}).

% during PO LEFT
Just before peak POF occurs in the observation phase, many changes were seen relative to baseline for both groups. The left hip angle has a small 3\% increase for the VF-S group, while the VF-R group has a 3\% decrease. Both changes occur from heel-strike until POF initiation. For the rigid group, this was coupled with a decrease in lateral gastrocnemius activity of 20\% through the stance phase leading up to peak POF, with the low stiffness group maintaining the same level of activity compared to baseline ($p < 0.01$). 

% during PO RIGHT
Changes to the contralateral leg were also observed. The VF-S group does have an increase in 10\% for the right biceps femoris at POF initiation, while the VF-R group has a rather large decrease in peak activation of 41\% ($p < 0.001$). For the right leg, the VF-S group has an increase in ankle plantarflexion by 9\%, while the VF-R group shows no change ($p < 0.05$). At the same time, subsequent changes were seen for the right knee extension, with the low stiffness group having a small increase of 5\%, and the rigid group having a small decrease of 4\% ($p < 0.05$). 

% after PO
Immediately following maximal POF initiation in the observation phase, several notable changes were seen. For the left leg, a small increase of 2\% was seen in knee flexion of the VF-S group, while the VF-R group at the same time frame has a 5\% decrease. The left leg also has a decrease of 9\% ($p < 0.05$) peak hip extension for the VF-R group, with the VF-S group seeing sustained peak hip extension. 

For the right leg, an increase in the second peak of ankle plantarflexion of 9\% ($p < 0.05$) during maximum POF for the low stiffness group is observed, while the rigid group has no increase relative to baseline. Muscle activity for the right leg during this period shows an increase of 9\% ($p < 0.05$) for the VF-S group biceps femoris, while the VF-R group has a 19\% ($p < 0.05$) decrease. This was coupled with a 22\% ($p < 0.05$) increase of rectus femoris of the right leg in the VF-S group, with only a 3\% increase in the VF-R group. At the same time as these changes, the right tibialis anterior has a remarkable increase of 18\% ($p < 0.01$) during right swing phase of only the low stiffness group, and the rigid group decreases activity during this period by 13\% ($p < 0.05$). 

% during left swing phase after PO
Immediately following left toe-off of the observation phase, the VF-R group has a small decrease in peak knee flexion of 2-5\%, while the VF-S group maintains peak knee flexion relative to baseline. Coupled with this change were several notable changes in left leg muscle activity. The vastus medialis decreases by 35\% ($p < 0.05$) in the VF-R group, and the VF-S group maintains muscle activity. Both of the hamstrings muscles have an average increase of 30\% ($p < 0.001$) for the VF-S group, while the VF-R group decreases by 29\% ($p < 0.05$) in average hamstring muscle activity. Also notable is the left tibialis anterior has an increase of 12\% ($p < 0.05$) for each peak activity during left swing phase for the VF-S group, with the VF-R group decreasing activity by 40\% ($p < 0.005$). The right leg for the VF-S group has a small increase in 4\% for peak hip extension and 2-6\% for the VF-R group.

\section{Discussion}

% training
%\Pd{Results from the training half of the experiment verify that both groups are trained to a similar level in terms of POF. All participants are able to successfully increase POF to meet or exceed the target at least 50\% of the time during the last section of the training half. It should also be noted that after the training half before participants are split into the respective group, there are no significant differences found in POF between groups. Training Acclimation phase duration was chosen in order to familiarize participants with the desired gait cadence, aligning with previous works using low-stiffness protocols on VST} \citep{yumbla2019,drolet2020,vc2021,vc2022,vc2023model}.\Pc{does not add anything. not needed}

% kinetics pof
%\Pd{Findings from POF data support the use of interventions increasing POF with visual feedback; however, lowering the stiffness during such an intervention seems to amplify these results. Because all groups are successful in increasing POF in both the training half and the trial half, the magnitude of the effect of the low stiffness was even more meaningful. While retention effects are realized regardless of having low stiffness or rigid terrain, the highest after-effect in terms of POF was in the VF-S group, showing significant promise for coupling visual bio-feedback and low stiffness in POF-based interventions. }\Pc{does not add anything. not needed}

% kinetics grf LEFT

\subsection{Ground Reaction Force Adaptations and Bilateral Loading}

The stance-phase duration increase of the left leg correlates with previous work \citep{vf1}. This finding is a direct side-effect of the strategies used when intentionally increasing POF. For analysis of the GRF profile, the most notable change in the VF-S group occurs with the POF having increased from baseline on the left leg (see Fig. \ref{fig:grf}). This change correlates with the rest of the VF-S group, with some other participants having larger after-effect increases in POF and some having lower increases. Also noted was a lowered peak weight-acceptance force that was shifted later in the gait cycle, which was not seen in the rigid group. In fact, the rigid group had an increase in weight acceptance force with a magnitude similar to the increase in peak POF seen in the low stiffness group. Both groups had very similar peak weight acceptance forces for baseline phase of the left leg. This observed adaptation, along with the expected increase in stance-phase duration, a known correlate of intentional POF increase \citep{vf1}, points toward the compliant environment facilitating a more natural and mechanically efficient gait strategy. This ability to decouple propulsive gains from high-impact landings is a key finding that supports the therapeutic potential of compliant-surface training.

%\Pc{although you report results here, what is missing is a discussion, of why this finding is important. Also, some of the findings are repeated here. This is the place for discussing the result, not repeating them}

% kinetics grf RIGHT
For the right leg, vertical GRF modestly decreased in the VF-S group and slightly increased in the VF-R group, suggesting a potential compensatory weight shift resulting from the intervention. This shift may explain the observed increases in left leg GRF and points to a meaningful after-effect: overall weight redistribution. Given that stroke survivors often exhibit reduced vertical GRF on the paretic leg, these results are especially relevant, highlighting the potential of this approach to promote more balanced loading and improve gait symmetry in stroke rehabilitation.

\subsection{Stance Phase Strategies: Joint Stiffness and Muscle Activation}
Findings from the preparation leading up to increasing POF indicate that the participants that received low-stiffness training allowed themselves to lower their center of mass by decreasing these specific joint angles, which provides evidence that the span between the hip and ankle was reduced in these participants. This finding is notable, as it indicates kinetically for the first time that joint stiffness was reduced as an after-effect of only low-stiffness POF training, and was likely a rebound effect of having to increase joint stiffness to overcome the compliant terrain. Also corroborating these results was the marked decrease in the peak weight acceptance force for the low stiffness group, further strengthening this claim.% \Pc{Give numbers and quantitative data to support this finding and claim, and relate to related literature. Otherwise, remove this paragraph}

%Because the rectus femoris is responsible for both knee extension and hip flexion, and the knee was allowed more flexion in this stage of the gait cycle for the VF-S group, the large increase in this muscle's activity can be seen as either a response to increasing knee stability, an allowance of the forward lean for the torso, or both. In stroke survivors, it is common to observe increases in joint stiffness after heel-strike \citep{li2021}, so if this after-effect translates to the patient population, it could be possible to achieve joint stiffness closer to baseline with this intervention. \Pc{this is a much stronger paragraph. Include some numbers through (even though the numbers appear in the Results section, and make a stronger case for application to stroke. I.e.. ``it could be possible to achieve joint stiffness closer to baseline with this intervention'' is not obvious this is a desired effect that leads to better functional outcomes in paretic gait. Why bringing joint stiffness to normal levels is important?}

The VF-S group exhibited a substantial increase in rectus femoris (RF) activity, which coincided with notably greater knee flexion during the loading response. Given the RF's biarticular role in knee extension and hip flexion, this elevated activity likely serves a dual purpose: providing dynamic knee stability to manage the compliant surface and controlling the forward torso lean required for higher propulsion. This finding has direct and significant implications for stroke rehabilitation. Many stroke survivors develop pathological joint stiffness, sucha as a ``stiff-knee gait,'' as a maladaptive compensatory strategy for muscle weakness and instability \citep{li2021}. This excessive rigidity is detrimental: it severely limits shock absorption, prevents a normal loading response, and blocks the kinematic adjustments needed for efficient propulsion. Our intervention appears to directly challenge this pathological pattern. By promoting increased knee flexion, the compliant surface discouraged a simple, rigid stabilization strategy. The corresponding surge in RF activity is therefore not pathological stiffening; rather, it represents a functional, learned muscular response to actively control a more dynamic and flexible joint. This is a critical distinction, suggesting a shift from a passive, locked-joint strategy to an active, controlled one. If this after-effect translates to paretic gait, it represents a potential method for re-training a more normative loading response, a crucial prerequisite for restoring shock absorption and improving propulsive power.

% during PO LEFT
During push-off, the left leg kinematic changes were small, but sustained throughout the stance phase. It cannot be concluded that the rectus femoris increases seen early in this period have the intention of increasing hip flexion in the group that underwent the compliant terrain. Rather, the small increase in hip flexion strengthens the point that knee joint stiffness was the likely target. The relative decrease in lateral gastrocnemius activity for the rigid group compared to the low-stiffness group suggests the present intervention for this muscle was critical, since the gastrocnemius plays a large role during the increased POF strategy of the intervention phase. It is apparent that decreased activity compared to baseline is seen as an effect of intentional POF increasing protocols in the rigid case, however this effect seems to disappear through the addition of lowered stiffness during the intervention. %\Pc{same comment as in the previous paragraph. add numbers, say why these effects are important/promising for stroke. Also, as noted in the beginning of the Discussion section, the paragraphs that remain should be included in subsections that have a clear and concise message. E.g., a possible title for a subsection is ``Effect of compliant surface on POF'' or similar}

\subsection{Swing Phase Mechanics and Bilateral Coordination}
% during PO RIGHT
At peak push-off of the observation phase, it was found that adding a compliant terrain while increasing POF in the intervention phase mitigates the decrease in peak hamstring activation of the contralateral (right) leg, while increasing ankle plantarflexion. Based on these results of the right leg, the further extended knee and pointed foot indicate that one potential after-effect of this study is an increase in target step length of the opposing leg. This finding nicely fits into known interventional strategies to counteract asymmetric step length in post-stroke individuals. At peak push off, the paretic leg has marked decrease in POF, with the contralateral leg having a corresponding decrease in step length. An increase in target step length directly counteracts the deficiency, while being trained to increase POF. This effect was only seen in groups that experienced walking under low stiffness conditions. Further research is needed to determine if step length asymmetry overall can be reduced with this intervention. %\Pc{better paragraph, add numbers and references}

% after PO
While small, the increase in knee flexion after push-off, at the initiation of swing phase, is a promising finding as it may have possible implications for an increase of toe clearance as an after-effect of the intervention. A decrease in toe clearance was seen in individuals post-stroke, so further research is needed to ascertain if an increase in toe clearance brings additional merit to this protocol. While both groups have an observed increase in POF compared to baseline, the decrease in POF for the group that does not experience low stiffness could be partially explained by the decrease in peak hip extension following peak push-off, connecting both the GRF and POF kinetic results with the kinematic observations.%\Pc{better paragraph, add numbers} 

%The co-contraction found in the upper muscles of the right leg indicates that the stability of the hip and knee joints was likely increased from baseline due to the contralateral intervention, but because both the rectus femoris and biceps femoris are biarticular muscles, it remains to be seen if the stability demands are increased more for the hip or the knee joint.\Pc{weak paragraph. not a clear message. It could be omitted.} 

Particularly, the large right tibialis anterior activity increase is exciting because increased tibialis anterior activation during swing phase is a target of many interventions focusing on decreasing the presence of drop-foot and increasing toe clearance. Walking under unilaterally compliant terrain has been shown to increase tibialis anterior activity of the contralateral leg during the intervention \citep{skidmore2016effect,skidmore2015}. This study shows for the first time that adding an intentionally increasing POF session coupled with unilaterally low stiffness walking not only shows an increase in contralateral tibialis anterior activity, but retains this as an after-effect of the intervention. Because this increase in activity has been shown in patients post-stroke \citep{skidmore2017}, this finding provides further evidence that this protocol has merit for translation to gait rehabilitation post-stroke, using the VST 2.%\Pc{MUCH better paragraph, add numbers, group in a subsection.}
% \Bd{The present result further solidifies the rationale that the after-effects mentioned in this work can translate to post-stroke gait rehabilitation using the VST 2, since previous results have already been shown in patients with stroke during the intervention with previous studies \citep{skidmore2017}.}

The finding of left tibialis anterior activity increase in the targeted side for the intervention is promising, since it may indicate that not only can ipsilateral stance phase kinetics be increased through adding low stiffness walking, but potentially swing phase toe clearance as well, broadening the scope of such a protocol to other pathologies known in the gait of individuals post-stroke. Combined together, the results for the VF-S group show a synergistic relationship between the left tibialis anterior and both left hamstrings in maintaining leg clearance during left swing phase. Because the muscle activity is significantly altered compared to baseline for the low stiffness group, but the increase in peak knee flexion is relatively minor, further research is needed to understand the potential outcomes of the increase in muscle activity. %\Pc{good paragraph, add numbers and references, group in a subsection.}

\subsection{Study Limitations and Future Work}

The immediate limitation of this work is its use of a healthy cohort, leaving the translation of these outcomes to clinical populations as an open question. While our specific findings on visual-compliance training are novel, the research pathway from healthy participants to patients is well-defined. Studies on split-belt adaptation, for example, have consistently shown that results from healthy cohorts can translate successfully to patient populations \citep{reisman2005,reisman2007}. Encouragingly, some studies have even found greater rehabilitative effects in patients than were observed in healthy participants \citep{reisman2009split,herrin2024}. Consequently, the clear future direction is to apply the present protocol to target populations (e.g., stroke survivors) to investigate its therapeutic potential.

Another limitation of this study is that our kinetic analysis was restricted to the vertical ground reaction force (GRF). Due to the constraints of our piezo-electric sensing equipment, we did not capture the anterior/posterior (A/P) forces, which are critical for a comprehensive understanding of walking kinetics and propulsion. Future work must utilize 3-axis load cells to measure the complete GRF profile. This enhanced instrumentation will allow for a direct analysis of propulsive A/P forces and enable the tracking of the center of pressure, providing deeper insight into the strategies used to increase POF.

Moreover, this study did not directly quantify knee joint stiffness, representing a key limitation. Our interpretations are based on indirect muscle activity patterns, such as the observed decrease in activity for both knee flexors and extensors (e.g., vastus medialis) during the swing phase in the rigid group. This preliminary finding suggests a potential decrease in swing-phase stiffness for that group. This is notable as it contrasts with the increase in stance-phase stiffness inferred for the compliant (VF-S) group. These opposing, and currently speculative, findings highlight the need for future work to use inverse dynamics to directly calculate joint stiffness. This will be essential to confirm and understand how these two training methods differentially modulate stiffness across the gait cycle.

\section{Conclusion}
This study investigated the after-effects of an intervention using visual force biofeedback with intentionally increasing push-off force (POF) while walking in a compliant environment. Including low-stiffness terrain as a tool for robot-assisted stroke therapy can improve subsequent gait outcome measures relative to walking on rigid terrain. The current study finds walking on unilateral low-stiffness environments such as the one emulated by VST 2 can have lasting increases in the activity of several muscles and joint angles during key moments in the gait cycle after the intervention when compared to baseline. These changes occur before, during, and after the push-off of both legs. Using these findings, rehabilitation protocols targeting the subsequent changes can be relevant to similar therapy techniques. This study shows for the first time a relationship between kinetic changes in the walking surface and relevant kinetic and kinematic changes in gait outcome measures that have benefits outside the current body of evidence, especially when used in conjunction with VST 2. The inclusion of modeling direct joint kinetics and anterior ground reaction forces should be the primary focus of future work, along with research directly studying the population of affected individuals who can benefit from such strategies presented in this work.

\balance
\bibliographystyle{apalike}
\bibliography{refs}

@article{chambershobbsVST2,
  title={The Variable Stiffness Treadmill 2: Development and Validation of a Unique Tool to Investigate Locomotion on Compliant Terrains},
  author={Chambers, Vaughn and Hobbs, Bradley and Gaither, William and Zhou, Anthony and Karakasis, Chrysostomos and Artemiadis, Panagiotis},
  journal={Journal of Mechanisms and Robotics},
  volume={17},
  number={3},
  year={2025},
  publisher={American Society of Mechanical Engineers Digital Collection}
}

@inproceedings{vf1,
  title={Intentional Increases in Push-off Force Coupled With Visual Feedback: Towards New Strategies in Robot-Assisted Gait Rehabilitation},
  author={Hobbs, Bradley and Artemiadis, Panagiotis},
  journal={Journal of Dynamic Systems, Measurement, and Control},
  booktitle={2024 IEEE RAS/EMBS International Conference on Biomedical Robotics and Biomechatronics (BioRob)},
  pages={1--6},
  year={2024},
  organization={IEEE}
}

@article{ck2024grf,
  title={An Energy-Based Framework for Robust Dynamic Bipedal Walking Over Compliant Terrain},
  author={Karakasis, Chrysostomos and Poulakakis, Ioannis and Artemiadis, Panagiotis},
  journal={Journal of Dynamic Systems, Measurement, and Control},
  volume={146},
  number={2},
  pages={021008},
  year={2024},
  publisher={American Society of Mechanical Engineers}
}

@inproceedings{ck2023pros,
  title={Adjusting the Quasi-Stiffness of an Ankle-Foot Prosthesis Improves Walking Stability during Locomotion over Compliant Terrain},
  author={Karakasis, Chrysostomos and Salati, Robert and Artemiadis, Panagiotis},
  booktitle={2023 IEEE/RSJ International Conference on Intelligent Robots and Systems (IROS)},
  pages={2140--2145},
  year={2023},
  organization={IEEE}
}

@article{ck2021real,
  title = {Real-time kinematic-based detection of foot-strike during walking},
  author = {Karakasis, Chrysostomos and Artemiadis, Panagiotis},
  year = {2021},
  journal = {Journal of Biomechanics},
  volume = {129},
  issn = {110849},
  copyright = {All rights reserved},
  pmid = {33844630}
}

@inproceedings{ck2021off,
  title = {{F-VESPA}: A kinematic-based algorithm for real-time heel-strike detection during walking},
  booktitle = {IEEE/RSJ International Conference on Intelligent Robots and Systems ({{IROS}})},
  author = {Karakasis, Chrysostomos and Artemiadis, Panagiotis},
  year = {2021},
  pages = {5098-5103},
  publisher = {{IEEE}},
  copyright = {All rights reserved}
}

@article{vc2023aftereffects,
  title={Using robot-assisted stiffness perturbations to evoke aftereffects useful to post-stroke gait rehabilitation},
  author={Chambers, Vaughn and Artemiadis, Panagiotis},
  journal={Frontiers in Robotics and AI},
  volume={9},
  pages={1073746},
  year={2023},
  publisher={Frontiers}
}

@article{franz2014,
  title={Real-time feedback enhances forward propulsion during walking in old adults},
  author={Franz, Jason R and Maletis, Michela and Kram, Rodger},
  journal={Clinical biomechanics},
  volume={29},
  number={1},
  pages={68--74},
  year={2014},
  publisher={Elsevier}
}

@article{spencer2021,
  title={Biofeedback for post-stroke gait retraining: a review of current evidence and future research directions in the context of emerging technologies},
  author={Spencer, Jacob and Wolf, Steven L and Kesar, Trisha M},
  journal={Frontiers in Neurology},
  volume={12},
  pages={637199},
  year={2021},
  publisher={Frontiers Media SA}
}

@article{genthe2018,
  title={Effects of real-time gait biofeedback on paretic propulsion and gait biomechanics in individuals post-stroke},
  author={Genthe, Katlin and Schenck, Christopher and Eicholtz, Steven and Zajac-Cox, Laura and Wolf, Steven and Kesar, Trisha M},
  journal={Topics in stroke rehabilitation},
  volume={25},
  number={3},
  pages={186--193},
  year={2018},
  publisher={Taylor \& Francis}
}

@article{schenck2017,
  title={Effects of unilateral real-time biofeedback on propulsive forces during gait},
  author={Schenck, Christopher and Kesar, Trisha M},
  journal={Journal of neuroengineering and rehabilitation},
  volume={14},
  pages={1--10},
  year={2017},
  publisher={Springer}
}

@book{perry1992,
  title={Gait Analysis: Normal and Pathological Function},
  author={Perry, Jacquelin and Burnfield, Judith},
  publisher={SLACK Incorporated},
  year = {1992},
  googlebooks = {hEdsAAAAMAAJ},
  isbn = {978-1-55642-192-1},
  langid = {english}
}

@article{balasubramanian2007,
  title={Relationship between step length asymmetry and walking performance in subjects with chronic hemiparesis},
  author={Balasubramanian, Chitralakshmi K and Bowden, Mark G and Neptune, Richard R and Kautz, Steven A},
  journal={Archives of physical medicine and rehabilitation},
  volume={88},
  number={1},
  pages={43--49},
  year={2007},
  publisher={Elsevier}
}

@inproceedings{hobbs2022,
  title = {A Systematic Method for Outlier Detection in Human Gait Data},
  booktitle = {{{IEEE 17th International Conference}} on {{Rehabilitation Robotics}} ({{ICORR}})},
  author = {Hobbs, Bradley and Artemiadis, Panagiotis},
  year = {2022},
  publisher = {{IEEE}},
  copyright = {All rights reserved}
}

@inproceedings{barkan2014,
  title={Variable stiffness treadmill ({VST}): A novel tool for the investigation of gait},
  author={Barkan, Andrew and Skidmore, Jeffrey and Artemiadis, Panagiotis},
  booktitle={2014 IEEE International Conference on Robotics and Automation (ICRA)},
  pages={2838--2843},
  year={2014},
  organization={IEEE}
}

@article{skidmore2016effect,
  title={On the effect of walking surface stiffness on inter-limb coordination in human walking: toward bilaterally informed robotic gait rehabilitation},
  author={Skidmore, Jeffrey and Artemiadis, Panagiotis},
  journal={Journal of neuroengineering and rehabilitation},
  volume={13},
  number={1},
  pages={1--11},
  year={2016},
  publisher={Springer}
}

@article{skidmore2017,
  title={Unilateral changes in walking surface compliance evoke dorsiflexion in paretic leg of impaired walkers},
  author={Skidmore, Jeffrey and Artemiadis, Panagiotis},
  journal={Journal of rehabilitation and assistive technologies engineering},
  volume={4},
  pages={2055668317738469},
  year={2017},
  publisher={SAGE Publications Sage UK: London, England}
}

@article{verma2012,
  title={Understanding gait control in post-stroke: implications for management},
  author={Verma, Rajesh and Arya, Kamal Narayan and Sharma, Pawan and Garg, RK},
  journal={Journal of bodywork and movement therapies},
  volume={16},
  number={1},
  pages={14--21},
  year={2012},
  publisher={Elsevier}
}

@article{banala2010novel,
  title={Novel gait adaptation and neuromotor training results using an active leg exoskeleton},
  author={Banala, Sai K and Agrawal, Sunil K and Kim, Seok Hun and Scholz, John P},
  journal={IEEE/ASME Transactions on mechatronics},
  volume={15},
  number={2},
  pages={216--225},
  year={2010},
  publisher={IEEE}
}

@inproceedings{banala2007,
  title={Active Leg Exoskeleton (ALEX) for gait rehabilitation of motor-impaired patients},
  author={Banala, Sai K and Agrawal, Suni K and Scholz, John P},
  booktitle={2007 IEEE 10th international conference on rehabilitation robotics},
  pages={401--407},
  year={2007},
  organization={IEEE}
}

@article{lemoyne2008,
  title={Virtual proprioception},
  author={Lemoyne, Robert and Coroian, Cristian and Mastroianni, Timothy and Grundfest, Warren},
  journal={Journal of Mechanics in Medicine and Biology},
  volume={8},
  number={03},
  pages={317--338},
  year={2008},
  publisher={World Scientific}
}

@article{banz2008,
  title={Computerized visual feedback: an adjunct to robotic-assisted gait training},
  author={Banz, Raphael and Bolliger, Marc and Colombo, Gery and Dietz, Volker and L{\"u}nenburger, Lars},
  journal={Physical therapy},
  volume={88},
  number={10},
  pages={1135--1145},
  year={2008},
  publisher={Oxford University Press}
}

@article{dwyer2010,
  title={Comparison of lower extremity kinematics and hip muscle activation during rehabilitation tasks between sexes},
  author={Dwyer, Maureen K and Boudreau, Samantha N and Mattacola, Carl G and Uhl, Timothy L and Lattermann, Christian},
  journal={Journal of athletic training},
  volume={45},
  number={2},
  pages={181--190},
  year={2010},
  publisher={The National Athletic Trainers' Association, Inc c/o Hughston Sports~…}
}

@article{reiman2012,
  title={A literature review of studies evaluating gluteus maximus and gluteus medius activation during rehabilitation exercises},
  author={Reiman, Michael P and Bolgla, Lori A and Loudon, Janice K},
  journal={Physiotherapy theory and practice},
  volume={28},
  number={4},
  pages={257--268},
  year={2012},
  publisher={Taylor \& Francis}
}

@inproceedings{akdougan2011,
  title={A muscular activation controlled rehabilitation robot system},
  author={Akdo{\u{g}}an, Erhan and {\c{S}}i{\c{s}}man, Zeynep},
  booktitle={Knowledge-Based and Intelligent Information and Engineering Systems: 15th International Conference, KES 2011, Kaiserslautern, Germany, September 12-14, 2011, Proceedings, Part I 15},
  pages={271--279},
  year={2011},
  organization={Springer Berlin Heidelberg}
}

@article{vallery2008,
  title={Reference trajectory generation for rehabilitation robots: complementary limb motion estimation},
  author={Vallery, Heike and Van Asseldonk, Edwin HF and Buss, Martin and Van Der Kooij, Herman},
  journal={IEEE transactions on neural systems and rehabilitation engineering},
  volume={17},
  number={1},
  pages={23--30},
  year={2008},
  publisher={IEEE}
}

@article{ji2008,
  title={Synthesis of a pattern generation mechanism for gait rehabilitation},
  author={Ji, Zhiming and Manna, Yazan},
  journal={Journal of Medical Devices},
  volume={2},
  number={3},
  year={2008},
  publisher={American Society of Mechanical Engineers Digital Collection}
}

@article{hesse1994,
  title={Gait outcome in ambulatory hemiparetic patients after a 4-week comprehensive rehabilitation program and prognostic factors.},
  author={Hesse, Stefan A and Jahnke, Matthias T and Bertelt, Christine M and Schreiner, Carl and L{\"u}cke, D and Mauritz, Karl-Heinz},
  journal={Stroke},
  volume={25},
  number={10},
  pages={1999--2004},
  year={1994},
  publisher={Am Heart Assoc}
}

@article{chen2003,
  title={Sagittal plane loading response during gait in different age groups and in people with knee osteoarthritis},
  author={Chen, Carl PC and Chen, Max JL and Pei, Yu-Cheng and Lew, Henry L and Wong, Pong-Yuen and Tang, Simon FT},
  journal={American journal of physical medicine \& rehabilitation},
  volume={82},
  number={4},
  pages={307--312},
  year={2003},
  publisher={LWW}
}

@article{kameyama1990,
  title={Electric discharge patterns of ankle muscles during the normal gait cycle.},
  author={Kameyama, O and Ogawa, R and Okamoto, T and Kumamoto, M},
  journal={Archives of physical medicine and rehabilitation},
  volume={71},
  number={12},
  pages={969--974},
  year={1990}
}

@article{dugan2005,
  title={Biomechanics and analysis of running gait},
  author={Dugan, Sheila A and Bhat, Krishna P},
  journal={Physical Medicine and Rehabilitation Clinics},
  volume={16},
  number={3},
  pages={603--621},
  year={2005},
  publisher={Elsevier}
}

@inproceedings{skidmore2015,
  title={Leg muscle activation Evoked by floor stiffness perturbations: A novel approach to robot-assisted gait rehabilitation},
  author={Skidmore, Jeffrey and Artemiadis, Panagiotis},
  booktitle={2015 IEEE International Conference on Robotics and Automation (ICRA)},
  pages={6463--6468},
  year={2015},
  organization={IEEE}
}

@article{mirelman2010,
  title={Effects of virtual reality training on gait biomechanics of individuals post-stroke},
  author={Mirelman, Anat and Patritti, Benjamin L and Bonato, Paolo and Deutsch, Judith E},
  journal={Gait \& posture},
  volume={31},
  number={4},
  pages={433--437},
  year={2010},
  publisher={Elsevier}
}

@article{chen2007,
  title={Ground reaction force patterns in stroke patients with various degrees of motor recovery determined by plantar dynamic analysis},
  author={Chen, C and Hong, PW and Chen, C and Chou, Shih Wei and Wu, C and Cheng, P and Tang, F and Chen, H},
  journal={Chang Gung medical journal},
  volume={30},
  number={1},
  pages={62},
  year={2007}
}

@article{hsu2019,
  title={Use of pelvic corrective force with visual feedback improves paretic leg muscle activities and gait performance after stroke},
  author={Hsu, Chao-Jung and Kim, Janis and Roth, Elliot J and Rymer, William Z and Wu, Ming},
  journal={IEEE Transactions on Neural Systems and Rehabilitation Engineering},
  volume={27},
  number={12},
  pages={2353--2360},
  year={2019},
  publisher={IEEE}
}

@article{dingwell1996,
  title={Use of an instrumented treadmill for real-time gait symmetry evaluation and feedback in normal and trans-tibial amputee subjects},
  author={Dingwell, JB and Davis, BL and Frazder, DM},
  journal={Prosthetics and orthotics international},
  volume={20},
  number={2},
  pages={101--110},
  year={1996},
  publisher={Taylor \& Francis}
}

@inproceedings{anson2013,
  title={Visual feedback during treadmill walking improves balance for older adults: a preliminary report},
  author={Anson, Eric and Kiemel, Tim and Tippawan, O and Jeka, John and others},
  booktitle={2013 International Conference on Virtual Rehabilitation (ICVR)},
  pages={166--167},
  year={2013},
  organization={IEEE}
}

@article{herrero2021,
  title={Gradually learning to increase gait propulsion in young unimpaired adults},
  author={Herrero, Luciana and Franz, Jason R and Lewek, Michael D},
  journal={Human Movement Science},
  volume={75},
  pages={102745},
  year={2021},
  publisher={Elsevier}
}

@article{white1997,
  title={Real-time dynamic visual feedback for altering gait of individuals after hip replacement},
  author={White, Scott and Klavoon, William and Lifeso, Robert},
  journal={Gait \& Posture},
  volume={2},
  number={5},
  pages={174--175},
  year={1997}
}

@article{tuthill2018,
  title = {Proprioception},
  author = {Tuthill, John C and Azim, Eiman},
  year = {2018},
  journal = {Current Biology},
  volume = {28},
  number = {5},
  pages = {R194-R203},
  issn = {0960-9822},
  doi = {10.1016/j.cub.2018.01.064}
}

@article{rand2018,
  title = {Proprioception Deficits in Chronic Stroke{\textemdash}{{Upper}} Extremity Function and Daily Living},
  author = {Rand, Debbie},
  year = {2018},
  month = mar,
  journal = {PLOS ONE},
  volume = {13},
  number = {3},
  pages = {e0195043},
  publisher = {{Public Library of Science}}
}

@article{preusser2015,
  title = {The Perception of Touch and the Ventral Somatosensory Pathway},
  author = {Preusser, Sven and Thiel, Sabrina D and Rook, Carolin and Roggenhofer, Elisabeth and Kosatschek, Anna and Draganski, Bogdan and Blankenburg, Felix and Driver, Jon and Villringer, Arno and Pleger, Burkhard},
  year = {2015},
  month = mar,
  journal = {Brain},
  volume = {138},
  number = {3},
  pages = {540--548},
  issn = {0006-8950},
  doi = {10.1093/brain/awu370}
}

@article{handelzalts2019,
  title = {Analysis of Brain Lesion Impact on Balance and Gait Following Stroke},
  author = {Handelzalts, Shirley and Melzer, Itshak and Soroker, Nachum},
  year = {2019},
  month = feb,
  journal = {Frontiers in Human Neuroscience},
  volume = {13},
  publisher = {{Frontiers Media S.A.}},
  issn = {16625161},
  doi = {10.3389/fnhum.2019.00149},
  urldate = {2020-08-12}
}

@inproceedings{skidmore2016a,
  title = {Sudden Changes in Walking Surface Compliance Evoke Contralateral {{EMG}} in a Hemiparetic Walker: {{A}} Case Study of Inter-Leg Coordination after Neurological Injury},
  booktitle = {2016 38th {{Annual International Conference}} of the {{IEEE Engineering}} in {{Medicine}} and {{Biology Society}} ({{EMBC}})},
  author = {Skidmore, Jeffrey and Artemiadis, Panagiotis},
  year = {2016},
  month = aug,
  pages = {4682--4685},
  publisher = {{IEEE}},
  issn = {1557170X},
  doi = {10.1109/EMBC.2016.7591772}
}

@article{chambers2023,
  title = {Using Robot-Assisted Stiffness Perturbations to Evoke Aftereffects Useful to Post-Stroke Gait Rehabilitation},
  author = {Chambers, Vaughn and Artemiadis, Panagiotis},
  year = {2023},
  journal = {Frontiers in Robotics and AI},
  volume = {9},
  issn = {2296-9144},
  urldate = {2023-01-04},
  copyright = {All rights reserved}
}

@article{xie2021,
  title = {How {{Compliance}} of {{Surfaces Affects Ankle Moment}} and {{Stiffness Regulation During Walking}}},
  author = {Xie, Kaifan and Lyu, Yueling and Zhang, Xianyi and Song, Rong},
  year = {2021},
  journal = {Frontiers in Bioengineering and Biotechnology},
  volume = {9},
  number = {October},
  pages = {1--10},
  doi = {10.3389/fbioe.2021.726051}
}

@article{li2021,
  title={How well do commonly used co-contraction indices approximate lower limb joint stiffness trends during gait for individuals post-stroke?},
  author={Li, Geng and Shourijeh, Mohammad S and Ao, Di and Patten, Carolynn and Fregly, Benjamin J},
  journal={Frontiers in Bioengineering and Biotechnology},
  volume={8},
  pages={588908},
  year={2021},
  publisher={Frontiers Media SA}
}

@ARTICLE{David2024,
  author={David, Pinto-Fernández and David, Rodriguez-Cianca and C., Moreno Juan and Diego, Torricelli},
  journal={IEEE Journal of Biomedical and Health Informatics}, 
  title={Human Locomotion Databases: A Systematic Review}, 
  year={2024},
  volume={28},
  number={3},
  pages={1716-1729},
doi={https://10.1109/JBHI.2023.3311677}
}

@article{Burden2003,
title = {How should we normalize electromyograms obtained from healthy participants? What we have learned from over 25years of research},
journal = {Journal of Electromyography and Kinesiology},
volume = {20},
number = {6},
pages = {1023-1035},
year = {2010},
issn = {1050-6411},
author = {Adrian Burden},
doi = {https://doi.org/10.1016/j.jelekin.2010.07.004}
}

@article{reisman2009split,
  title={Split-belt treadmill adaptation transfers to overground walking in persons poststroke},
  author={Reisman, Darcy S and Wityk, Robert and Silver, Kenneth and Bastian, Amy J},
  journal={Neurorehabilitation and neural repair},
  volume={23},
  number={7},
  pages={735--744},
  year={2009},
  publisher={SAGE Publications Sage CA: Los Angeles, CA}
}

@article{reisman2005,
  title={Interlimb coordination during locomotion: what can be adapted and stored?},
  author={Reisman, Darcy S and Block, Hannah J and Bastian, Amy J},
  journal={Journal of neurophysiology},
  volume={94},
  number={4},
  pages={2403--2415},
  year={2005},
  publisher={American Physiological Society}
}

@article{reisman2007,
  title={Locomotor adaptation on a split-belt treadmill can improve walking symmetry post-stroke},
  author={Reisman, Darcy S and Wityk, Robert and Silver, Kenneth and Bastian, Amy J},
  journal={Brain},
  volume={130},
  number={7},
  pages={1861--1872},
  year={2007},
  publisher={Oxford University Press}
}

@article{franz2016,
  title={Visuomotor entrainment and the frequency-dependent response of walking balance to perturbations},
  author={Franz, Jason R and Francis, Carrie A and Allen, Matthew S and Thelen, Darryl G},
  journal={IEEE transactions on neural systems and rehabilitation engineering},
  volume={25},
  number={8},
  pages={1135--1142},
  year={2016},
  publisher={IEEE}
}

@article{ensink2025,
  title={Effect of feedback on foot strike angle and forward propulsion in people with stroke},
  author={Ensink, Carmen J and Hofstad, Cheriel J and Van Ee, Ren{\'a} and Keijsers, No{\"e}l LW},
  journal={IEEE Transactions on Neural Systems and Rehabilitation Engineering},
  year={2025},
  publisher={IEEE}
}

@article{kim2024gait,
  title={Gait Symmetric Adaptation and Aftereffect Through Concurrent Split-Belt Treadmill Walking and Explicit Visual Feedback Distortion},
  author={Kim, Seung-Jae and Save, Omik and Tanner, Emily and Marquez, Arianna and Lee, Hyunglae},
  journal={IEEE Transactions on Biomedical Engineering},
  year={2024},
  publisher={IEEE}
}

@article{kim2015,
  title={Effects of visual feedback distortion on gait adaptation: comparison of implicit visual distortion versus conscious modulation on retention of motor learning},
  author={Kim, Seung-Jae and Ogilvie, Mitchell and Shimabukuro, Nathan and Stewart, Trevor and Shin, Joon-Ho},
  journal={IEEE Transactions on Biomedical Engineering},
  volume={62},
  number={9},
  pages={2244--2250},
  year={2015},
  publisher={IEEE}
}

@article{awad2014,
  title={Targeting paretic propulsion to improve poststroke walking function: a preliminary study},
  author={Awad, Louis N and Reisman, Darcy S and Kesar, Trisha M and Binder-Macleod, Stuart A},
  journal={Archives of physical medicine and rehabilitation},
  volume={95},
  number={5},
  pages={840--848},
  year={2014},
  publisher={Elsevier}
}

@article{awad2020,
  title={These legs were made for propulsion: advancing the diagnosis and treatment of post-stroke propulsion deficits},
  author={Awad, Louis N and Lewek, Michael D and Kesar, Trisha M and Franz, Jason R and Bowden, Mark G},
  journal={Journal of NeuroEngineering and Rehabilitation},
  volume={17},
  pages={1--16},
  year={2020},
  publisher={Springer}
}

@article{krishnamoorthy2008,
  title={Gait training after stroke: a pilot study combining a gravity-balanced orthosis, functional electrical stimulation, and visual feedback},
  author={Krishnamoorthy, Vijaya and Hsu, Wei-Li and Kesar, Trisha M and Benoit, Daniel L and Banala, Sai K and Perumal, Ramu and Sangwan, Vivek and Binder-Macleod, Stuart A and Agrawal, Sunil K and Scholz, John P},
  journal={Journal of neurologic physical therapy},
  volume={32},
  number={4},
  pages={192--202},
  year={2008},
  publisher={LWW}
}

@article{liu2021,
  title={Comparison of the effects of real-time propulsive force versus limb angle gait biofeedback on gait biomechanics},
  author={Liu, Justin and Santucci, Vincent and Eicholtz, Steven and Kesar, Trisha M},
  journal={Gait \& posture},
  volume={83},
  pages={107--113},
  year={2021},
  publisher={Elsevier}
}

@article{ting2015,
  title={Neuromechanical principles underlying movement modularity and their implications for rehabilitation},
  author={Ting, Lena H and Chiel, Hillel J and Trumbower, Randy D and Allen, Jessica L and McKay, J Lucas and Hackney, Madeleine E and Kesar, Trisha M},
  journal={Neuron},
  volume={86},
  number={1},
  pages={38--54},
  year={2015},
  publisher={Elsevier}
}

@article{giggins2013,
  title={Biofeedback in rehabilitation},
  author={Giggins, Oonagh M and Persson, Ulrik McCarthy and Caulfield, Brian},
  journal={Journal of neuroengineering and rehabilitation},
  volume={10},
  pages={1--11},
  year={2013},
  publisher={Springer}
}

@article{sigrist2013,
  title={Augmented visual, auditory, haptic, and multimodal feedback in motor learning: a review},
  author={Sigrist, Roland and Rauter, Georg and Riener, Robert and Wolf, Peter},
  journal={Psychonomic bulletin \& review},
  volume={20},
  pages={21--53},
  year={2013},
  publisher={Springer}
}

@article{nelson2007,
  title={The role of biofeedback in stroke rehabilitation: past and future directions},
  author={Nelson, Lonnie A},
  journal={Topics in stroke rehabilitation},
  volume={14},
  number={4},
  pages={59--66},
  year={2007},
  publisher={Taylor \& Francis}
}

@article{hsiao2015,
  title={Mechanisms to increase propulsive force for individuals poststroke},
  author={Hsiao, HaoYuan and Knarr, Brian A and Higginson, Jill S and Binder-Macleod, Stuart A},
  journal={Journal of neuroengineering and rehabilitation},
  volume={12},
  pages={1--8},
  year={2015},
  publisher={Springer}
}

@article{roelker2019,
  title={Paretic propulsion as a measure of walking performance and functional motor recovery post-stroke: A review},
  author={Roelker, Sarah A and Bowden, Mark G and Kautz, Steven A and Neptune, Richard R},
  journal={Gait \& posture},
  volume={68},
  pages={6--14},
  year={2019},
  publisher={Elsevier}
}

@article{huang2006,
  title={Recent developments in biofeedback for neuromotor rehabilitation},
  author={Huang, He and Wolf, Steven L and He, Jiping},
  journal={Journal of neuroengineering and rehabilitation},
  volume={3},
  pages={1--12},
  year={2006},
  publisher={Springer}
}

@article{basmajian1981,
  title={Biofeedback in rehabilitation: a review of principles and practices.},
  author={Basmajian, JV},
  journal={Archives of physical medicine and rehabilitation},
  volume={62},
  number={10},
  pages={469--475},
  year={1981}
}

@article{richards2017,
  title={Gait retraining with real-time biofeedback to reduce knee adduction moment: systematic review of effects and methods used},
  author={Richards, Rosie and van den Noort, Josien C and Dekker, Joost and Harlaar, Jaap},
  journal={Archives of physical medicine and rehabilitation},
  volume={98},
  number={1},
  pages={137--150},
  year={2017},
  publisher={Elsevier}
}

@article{zelik2016,
  title={A unified perspective on ankle push-off in human walking},
  author={Zelik, Karl E and Adamczyk, Peter G},
  journal={Journal of Experimental Biology},
  volume={219},
  number={23},
  pages={3676--3683},
  year={2016},
  publisher={The Company of Biologists Ltd}
}

@article{lewek2012,
  title={Use of visual and proprioceptive feedback to improve gait speed and spatiotemporal symmetry following chronic stroke: a case series},
  author={Lewek, Michael D and Feasel, Jeff and Wentz, Erin and Brooks Jr, Frederick P and Whitton, Mary C},
  journal={Physical therapy},
  volume={92},
  number={5},
  pages={748--756},
  year={2012},
  publisher={Oxford University Press}
}

@article{schenck2019,
  title={Haptic biofeedback induces changes in ankle push-off during walking},
  author={Schenck, Christopher and Bakke, Duncan and Besier, Thor},
  journal={Gait \& posture},
  volume={74},
  pages={76--82},
  year={2019},
  publisher={Elsevier}
}

@article{herrin2024,
  title={Robotic Ankle Exoskeleton and Limb Angle Biofeedback for Assisting Stroke Gait: A Feasibility Study},
  author={Herrin, Kinsey R and Pan, Yi-Tsen and Kesar, Trisha M and Sawicki, Gregory S and Young, Aaron J},
  journal={IEEE Robotics and Automation Letters},
  year={2024},
  publisher={IEEE}
}

\end{document}